\documentclass{article}

\usepackage[final]{neurips_2023}

\usepackage[utf8]{inputenc} % allow utf-8 input
\usepackage[T1]{fontenc}    % use 8-bit T1 fonts
\usepackage{hyperref}       % hyperlinks
\usepackage{url}            % simple URL typesetting
\usepackage{booktabs}       % professional-quality tables
\usepackage{amsfonts}       % blackboard math symbols
\usepackage{nicefrac}       % compact symbols for 1/2, etc.
\usepackage{microtype}      % microtypography
\usepackage{algorithm}
\usepackage{amsmath,mathtools,physics}
\usepackage{algpseudocode}
\usepackage{color}
\usepackage{amstext}
\usepackage{stmaryrd}
\usepackage{enumerate}
\usepackage{pifont}
\usepackage{prettyref}
\usepackage{hyperref}

\usepackage{graphicx}
\usepackage{tabularx}
\usepackage{caption}
\usepackage{booktabs}
\usepackage{xcolor}
\usepackage{amsmath}

\makeatletter
\newcommand{\printfnsymbol}[1]{%
  \textsuperscript{\@fnsymbol{#1}}%
}
\makeatother

\title{Self-Supervised Image Captioning with CLIP}

\author{%
  Chuanyang Jin \\
  New York University\\
  \texttt{cj2133@nyu.edu} \\
}

\begin{document}

\maketitle

\begin{abstract}
Image captioning, a fundamental task in vision-language understanding, seeks to generate accurate natural language descriptions for provided images. Current image captioning approaches heavily rely on high-quality image-caption pairs, which can be hard to obtain for many domains. To address this, we introduce a self-supervised image captioning method. After learning an initial signal from a small labeled dataset, our method transitions to self-supervised learning on unlabeled data, leveraging the auxiliary task of enhancing the CLIP relevance between images and generated captions. Remarkably, despite utilizing less than 2\% of the labeled COCO dataset, our method delivers a performance comparable to state-of-the-art models trained on the complete dataset. Human evaluations further reveal that our method produces captions with greater distinctiveness and informativeness, two attributes inherently challenging to achieve through supervised learning.
\end{abstract}

\section{Introduction}
Image captioning aims to describe images with syntactically and semantically meaningful sentences, thereby bridging the gap between vision and language. Current approaches in this domain often rely on supervised learning, which poses certain challenges.

Firstly, acquiring image-caption pairs is a challenging endeavor for many domains, thus it is desirable to have a method that doesn't need much supervision. Despite this, prevailing methods rely heavily on extensive collections of captioned images for training. In some cases, these models even necessitate multiple reference captions and additional annotations \citep{stefanini2022show}.

Second, the quality of reference captions can be less than ideal. As will be shown later, current models can produce captions that match the quality of the reference captions in well-known datasets like COCO. Nevertheless, the reliance on reference-based similarity objectives prevents these models from surpassing this quality threshold. For example, models trained on text similarity objectives tend to overlook specific details that set one image apart from others, as public datasets' reference captions typically describe only the most conspicuous and common objects \citep{cho2022fine}.

In light of these challenges, our research introduces a two-stage self-supervised captioning method. Notably, our method (1) significantly reduces dependency on reference captions through a self-supervised mechanism, and (2) while leveraging insights from these captions, surpasses the quality of the original reference captions in certain evaluations.

\begin{figure}[t!]
    \centering
    \captionsetup{skip=2pt}
    \begin{minipage}{0.24\textwidth}
        \centering
        \includegraphics[width=\textwidth]{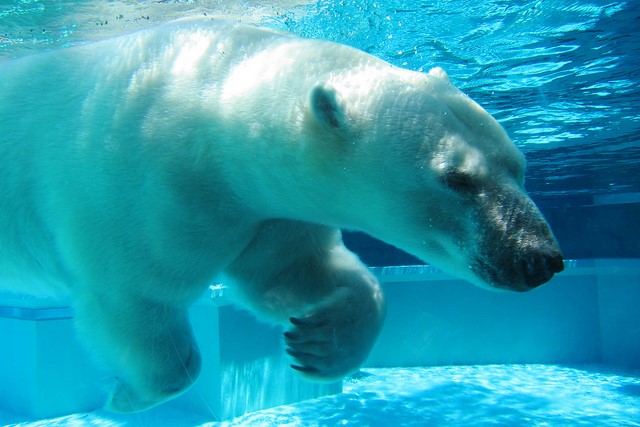}
        \caption*{\color{darkgray}\small A large polar bear swimming in a pool of water.}
        \label{fig:image1}
    \end{minipage}
    \hfill
    \begin{minipage}{0.24\textwidth}
        \centering
        \includegraphics[width=\textwidth]{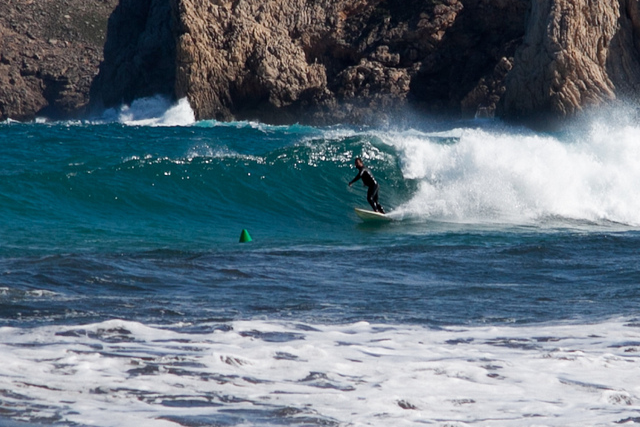}
        \caption*{\color{darkgray}\small A man riding a wave on top of a surfboard.}
        \label{fig:image2}
    \end{minipage}
    \hfill
    %\vspace{10pt} % Adjust this space to your need
    \begin{minipage}{0.24\textwidth}
        \centering
        \includegraphics[width=\textwidth]{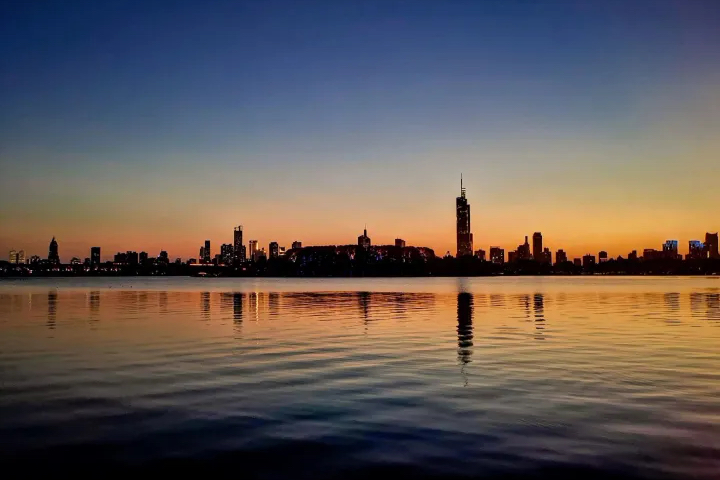}
        \caption*{\color{darkgray}\small A city skyline and a large body of water in the sunset.}
        \label{fig:image3}
    \end{minipage}
    \hfill
    \begin{minipage}{0.24\textwidth}
        \centering
        \includegraphics[width=\textwidth]{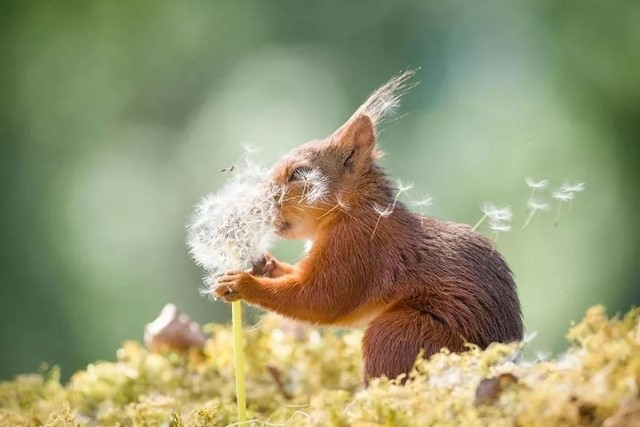}
        \caption*{\color{darkgray}\small A small brown squirrel standing on the grass.}
        \label{fig:image4}
    \end{minipage}
    \caption{Example captions generated by our method.}
    \label{fig:examples}
\end{figure}

\section{Method}
\subsection{Image Captioning with CLIP}
Large research efforts have been devoted to image captioning. Typically, an image captioning method first encodes an image into a visual representation, which is then decoded to produce the final captions. The visual encoder extracts the representation from either classification networks \citep{chen2017sca}  \citep{fang2015captions} \citep{xu2015show},  or object detection networks \citep{anderson2018bottom} \citep{li2020oscar} \citep{zhou2020unified}, with the latter providing more expressive features. Some models introduce a self-attention mechanism \citep{herdade2019image} or Vision Transformers \citep{dosovitskiy2020image} to make better use of visual cues. When it comes to the textual decoder, different models use LSTM variants \citep{chen2018regularizing} \citep{vinyals2015show} or transformer-based architectures \citep{herdade2019image} \citep{luo2021dual}. The emergence of large language models like GPT \citep{brown2020language} presents promising alternatives. %\citep{radford2018improving}\citep{radford2019language} %devoted to image captioning \citep{stefanini2022show}  classification networks \citep{chen2014learning} llm like BERT \cite{devlin2018bert} and 

Contrastive Language-Image Pre-training (CLIP) \citep{radford2021learning} 
leverages vision-language pre-training on abundant image-caption pairs and produces cross-modal contextual representations. ClipCap \citep{mokady2021clipcap} leverages CLIP representations for image captioning. First, it employs CLIP's visual encoder to generate an encoded image representation. Next, a mapping network—either a transformer or MLP—translates the CLIP embedding to the language model space, producing a prefix to the caption. Finally, the language model auto-regressively generates the predicted image captions. By leveraging multimodal understanding from existing models instead of learning new semantic entities, ClipCap produces state-of-the-art results with a simpler and faster model.

\subsection{Our Method}
We propose an enhancement to the ClipCap architecture to reduce its dependency on captioned images. This is achieved by introducing an auxiliary task that aims to maximize the matching, as determined by CLIP, between the image and the captions produced.

To integrate this auxiliary task as a direct replacement for the original training objective, we face a challenge during optimization: The language model generates captions by sequentially predicting each word token, which involves a probability distribution across the vocabulary for each token. The selection of the next token typically employs either greedy decoding or beam search, both relying on the Softmax operation. However, the discrete sampling operation following Softmax is non-differentiable, thus interrupting the flow of gradients. To overcome this challenge, we use the Gumbel-Softmax method \citep{jang2016categorical}, which provides a reparameterization trick for discrete variables. This technique allows us to transform the non-differentiable operation into a differentiable one, thereby enabling our gradient-based optimization on the auxiliary task.

Nonetheless, in the absence of initial signals, the generated text is purely random and incoherent. As a result, the image-caption relevance calculated would stay extremely low. This creates a significant challenge in further optimizing the model. Given that the search space for potential inputs is exceptionally vast and discrete, the loss will fluctuate wildly and fail to converge. To overcome this challenge, we introduce an initial supervised training stage before the self-supervised training stage. This supervised stage equips the model with an essential set of preliminary signals, thereby establishing a baseline proficiency that can be further refined during the self-supervised training stage. An ablation study highlighting the significance of this supervised stage can be found in Appendix~\ref{sec:more_results}.

Breaking down our methodology, during the supervised training stage, an initial signal is established using a small dataset of 10,000 captioned images, constituting less than 2$\%$ of the standard training division in the COCO-captions dataset \citep{lin2014microsoft}. For each image paired with its corresponding caption, the captions are considered as a sequence of tokens represented as $\boldsymbol{c} = c_1, ..., c_l$. Using the CLIP model, we generate the visual embedding $\boldsymbol{v}$. This is then transformed by our mapping network to produce a series of prefix embeddings $\boldsymbol{p} = p_1, \dots, p_k$, where each shares the same dimensionality as the word embedding. Following this, we concatenate the prefix embeddings with the caption embeddings, $c_1,\dots,c_{i-1}$, and feed the concatenation into the language model LLaMA 2 to predict the subsequent token $c_i$. We either train the mapping component or finetune the LLaMA 2 using a cross-entropy loss $\mathcal{L}_{\text{supv}} = -\Sigma_{i=1}^l \log p_\theta (c_i | p_1, \dots, p_k, c_1, \dots, c_{i-1})$.

In the self-supervised training stage, each step involves two main steps. First, we employ LLaMA 2 to generate a caption, $\boldsymbol{c}_{\text{model}}$, corresponding to a specific image. Subsequently, we train the model based on the relevance between the image and its generated caption. The loss function for this phase is crafted to be the inverse of the image-caption relevance as evaluated by the CLIP embeddings, represented as $\mathcal{L}_{\text{unsupv}} = -\cos(\boldsymbol{v}, \boldsymbol{c}_{\text{model}})$.

% Training over 10 epochs on this small dataset requires only a few minutes on a single GPU. 

%$$\mathcal{L}_{\text{supv}} = - \sum_{i=1}^l \log p_\theta(c_i | p_1, \dots, p_k, c_1, \dots, c_{i-1})$$

% The subsequent stage involves self-supervised training of the model. We develop a cohesive framework that integrates both stages, thereby enabling smooth transitions between stages throughout the training cycles.

% \subsection{Self-supervised Stage with CLIP Objective}

% Typically, this stage takes around 10 epochs of training, a process that can be completed in roughly 10 GPU hours.

 % Suppose we have an image with corresponding CLIP visual embedding $\boldsymbol{v}$ and a candidate caption with corresponding CLIP textual embedding $\boldsymbol{c}$, we employ cosine similarity to compute our loss function as:
%$$\mathcal{L}_{\text{unsupv}} = -\cos(\boldsymbol{v}, \boldsymbol{c})$$ 

\section{Experiments}
\subsection{Comprehensive Evaluation}
Current methods for evaluating the quality of generated text, like in image captioning, are mainly based on comparing the text to a set of reference texts. These methods include n-gram overlap techniques such as BLEU \citep{papineni2002bleu}, METEOR \citep{denkowski2014meteor}, and CIDEr \citep{vedantam2015cider}. There are also other metrics like SPICE \citep{anderson2016spice} and TIGEr \citep{jiang2019tiger} that go beyond simple overlap and incorporate more sophisticated models of similarity between reference and candidate texts. These metrics operate under the belief that the reference captions are of high quality and serve as a benchmark for the generated captions to reach. However, our subsequent analyses indicate that many captions created by models are actually better than reference captions, offering more precise descriptions. This suggests a need for new metrics that do not solely rely on how close the generated caption is to a reference caption.

Recent metrics suggest using relevance scores from Vision-Language Models (VLMs) like BERTScore \citep{zhang2019bertscore}, ViLBERTScore \citep{lee2020vilbertscore}, UMIC \citep{lee2021umic}, and CLIPScore \citep{hessel2021clipscore}. These VLMs are trained on large datasets and acquire a rich semantic understanding of the relationship between images and texts, enabling them evaluate captions in a zero-shot manner. Building on this concept, we introduce the RefCompare Score, which leverages CLIP alongside available reference captions. This metric offers a more consistent evaluation and delves deeper into understanding how a method measures up against the reference captions. In particular, the RefCompare Score calculates the average proportion of reference captions that score lower in CLIP relevance to the image than the generated caption does.

\begin{figure}[h]
    \centering
    \includegraphics[width=0.55\textwidth]
    {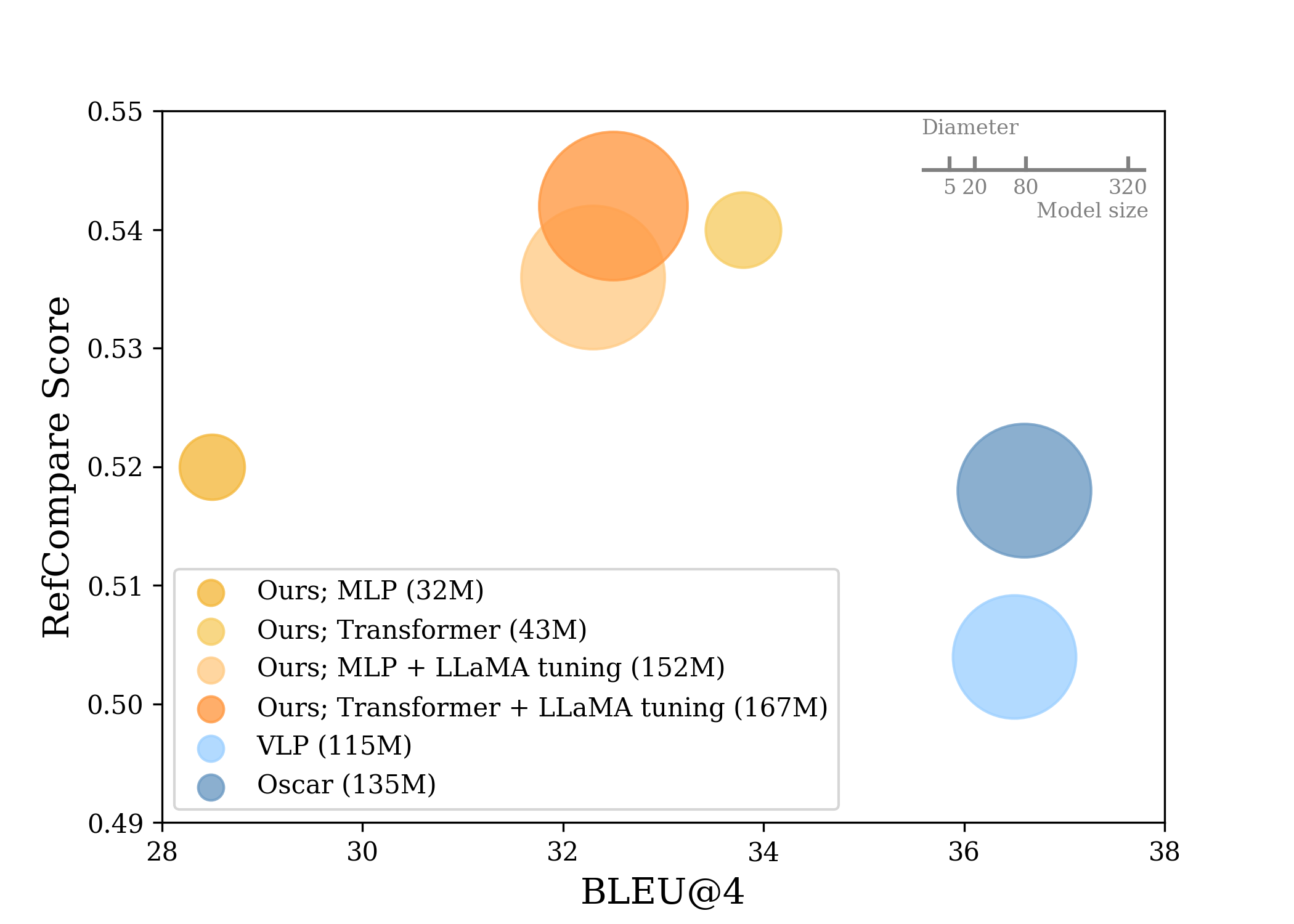}
    \caption{Comparative evaluation of two state-of-the-art models against four variants of our method.}
    \label{fig:results}
\end{figure}

In this context, a RefCompare Score equal to or higher than 0.5 suggests that the model's captions are of comparable or better quality than the benchmark captions. As illustrated in Figure~\ref{fig:results}, top-performing models have the ability to create captions that are not just similar, but often better in quality than the standard references. This underscores the importance of having evaluation metrics that measure beyond mere textual similarity.

For a comprehensive assessment, we use the traditional BLEU score along with our RefCompare Score to compare our baseline methods and proposed approach. Acknowledging that existing metrics might not completely reflect the subtleties of human judgment, we also conduct a human evaluation to measure two important aspects: distinctiveness and informativeness.

\subsection{Baselines and Variants of Our Method}
We evaluate the performance of state-of-the-art vision-language models, notably \textbf{Oscar} \citep{li2020oscar} and \textbf{VLP} \citep{zhou2020unified}, adhering to the finetuning and inference strategies described in their respective publications.

For our method, we experiment with four variants: two that involve training the mapping network with a frozen language model, and two that include the finetuning of the language model. The mapping network is either architected as an MLP or a Transformer network. We refer to these as \textbf{MLP}, \textbf{Transformer}, \textbf{MLP + LLaMA finetuning}, and \textbf{Transformer + LLaMA finetuning}.

\subsection{Results}

We compare our method with the Oscar and VLP baselines. It is noteworthy that our model has fewer parameters and needs less time to train. In terms of performance, while our model yields slightly lower BLEU@4 scores, similar to the supervised ClipCap method, it stands out by securing substantially higher RefCompare Scores.

Looking at the four variants of our method, our default Transformer configuration delivers the highest BLEU@4 score and excellent RefCompare Score, while maintaining a relatively small model size. The MLP configuration, on the other hand, presents weaker performance, likely due to its limitation in expressiveness. Both the MLP + LLaMA tuning and Transformer + LLaMA tuning versions yield decent results, but they do not distinctly outperform the Transformer, even with more trainable parameters and longer training times. This observation suggests that finetuning the language model in conjunction with using a Transformer mapping network could lead to an excess of expressive power.
 
\subsection{Human Evaluation}
We conduct a human evaluation to assess preferences between captions generated by the Oscar model and those produced by our method. We looked at two key aspects: how unique the captions are and how much information they convey. For this, we sample 1000 random images from the COCO-captions dataset and create captions using both methods. We then engage five independent evaluators to compare the distinctiveness and informativeness of each set of captions without disclosing which are generated by our model. The results show that our model’s captions are perceived as more distinctive in 58.6\% of the cases, and more informative in 69.2\% of the cases.

We present some examples in Appendix~\ref{sec:more_examples}. As we can see, supervised methods relying on references often employ more commonplace words in their descriptions, leading to a better match with reference captions. In contrast, our method generates more distinct, semantically rich, and ``human-like" interpretations. While these may not always echo the phrasing of the reference captions and might contain occasional errors, they largely resonate with human intuition and common sense. Notably, while both the human-annotated captions and those from reference-based methods tend to focus on describing the most prominent and frequently encountered objects, our method seeks to incorporate additional objects within the scene, providing a more comprehensive and detailed description.

\section{Conclusion}
We introduce a self-supervised image captioning method that leverages the auxiliary task of CLIP relevance. This method not only demonstrates remarkable performance in similarity-based metrics but also achieves state-of-the-art results in evaluations based on Vision-Language Models (VLM). Unlike existing methods, our method significantly diminishes the dependency on captioned images and generates captions that are more distinctive, informative, and aligned with human preference.

\section*{Acknowledgements}
We thank Peiqi Liu, He He, Saining Xie, Tianmin Shu, and Hao Zhu for their valuable insights shared during our discussions.
\bibliographystyle{plainnat}
\bibliography{reference}

\newpage
\appendix
\section{More Results} \label{sec:more_results}
\begin{figure}[h]
    \centering
    \includegraphics[width=0.55\textwidth]
    {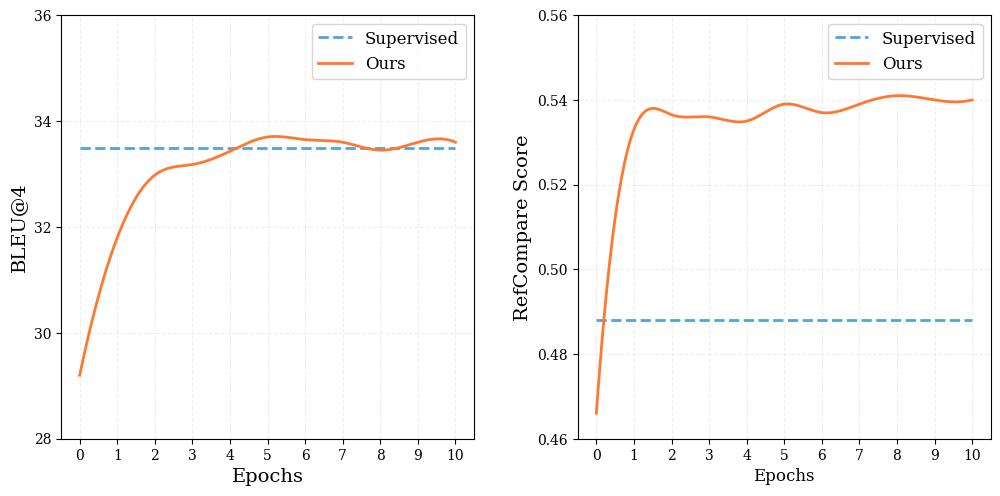}
    \caption{BLEU@4 and RefCompare Scores during the self-supervised stage.}
    \label{fig:scores}
\end{figure}

In Figure~\ref{fig:scores}, we present the model's performance in terms of BLEU@4 and RefCompare Scores during the self-supervised training stage. 
Initially, the performance trails that of the supervised-only approach, largely due to the limited size of the dataset used in the supervised phase. However, as training progresses, we observe a notable increase in performance. While there are occasional fluctuations, the general trajectory is upward, stabilizing after about 10 epochs. Impressively, our method not only narrowly exceeds the BLEU@4 score of the supervised method but also significantly outperforms it in the RefCompare Score.

\begin{figure}[h]
    \centering
    \includegraphics[width=0.55\textwidth]
    {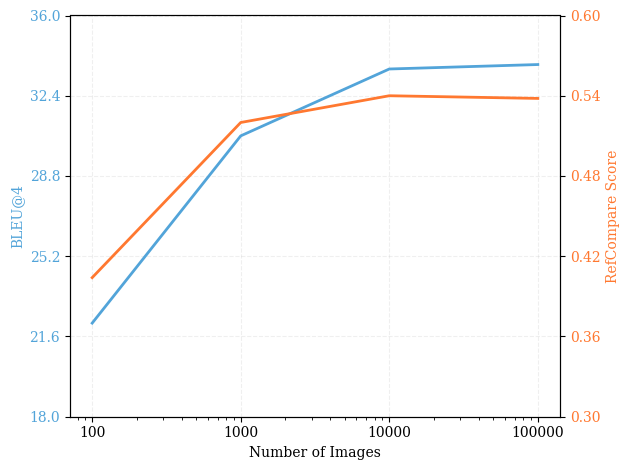}
    \caption{Final BLEU@4 and RefCompare Scores with the variation in the number of images used in the supervised stage.}
    \label{fig:scores_vs_num}
\end{figure}

In Figure~\ref{fig:scores_vs_num}, we depict the relationship between the size of the labeled training dataset used during the supervised stage and the model's final performance. Notably, even a modest dataset of 100 images for supervised training is sufficient to guide the model effectively through the subsequent self-supervised stage. As we increase the size of the supervised training set, the model's performance improves consistently. However, beyond an approximate limit of 10,000 images, adding more training data does not lead to a significant enhancement of the model's performance.

\newpage
\section{More Examples} \label{sec:more_examples}
\begin{figure*}[h]
%\small
\footnotesize
\begin{tabularx}{\textwidth}{lXXXXX}
& \includegraphics[width=\linewidth]{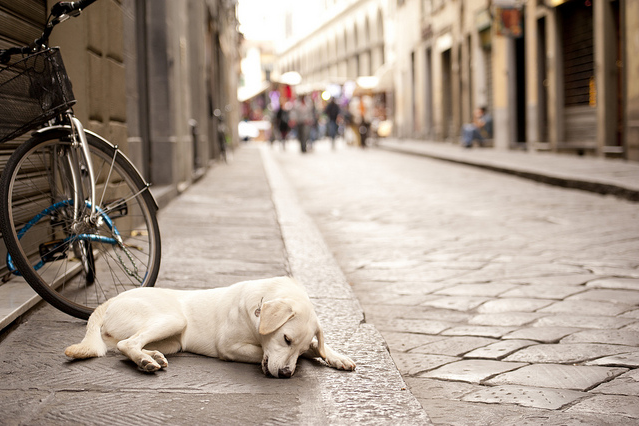} & \includegraphics[width=\linewidth]{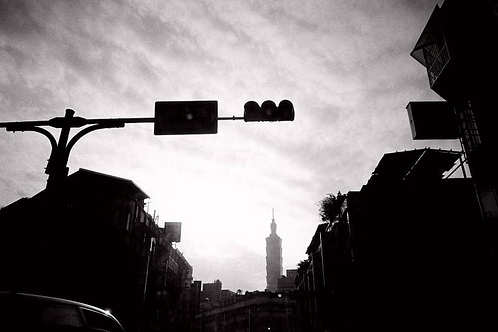} & \includegraphics[width=\linewidth]{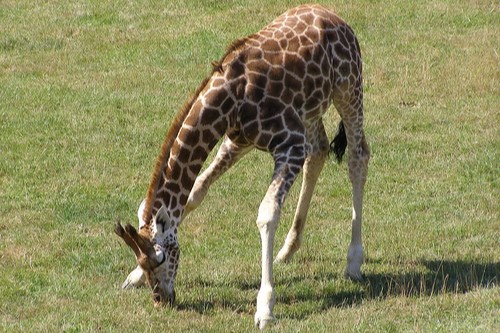} & \includegraphics[width=\linewidth]{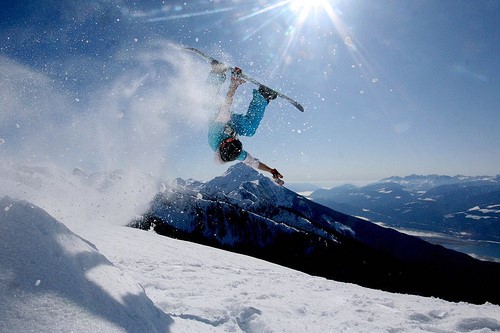} & \includegraphics[width=\linewidth]{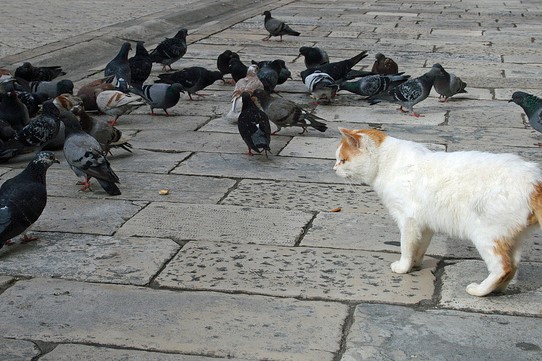} \\
\hline
Reference & 
A puppy rests on the street next to a bicycle & 
A traffic light and street sign surrounded by buildings & 
A giraffe bending over while standing on green grass &
A person up in the air, upside down while outside & 
A very cute cat near a bunch of birds \\
\hline
Supervised & 
A puppy sleeping on the street ($0.02$) & 
A city in the fog ($0.68$) & 
A giraffe that is standing in the grass ($0.04$) &
A man flying through the air while riding a snowboard ($1$) &
A white cat standing next to a bunch of pigeons ($0.98$) \\
\hline
Ours & 
A white dog laying on the street next to a bike ($0.15$) & 
A black and white photo of a traffic light ($0.99$) &
A giraffe eating grass in a grassy field ($0.16$) & 
A person on a snowboard jumping over a snow-covered slope ($1$) &
A cat approaching a bunch of pigeons ($0.97$) \\
\end{tabularx}
%\vspace{\baselineskip}
\caption{Uncurated captions generated through both the supervised approach and our method for five images in the COCO-captions dataset. RefCompare scores are shown in parentheses.}
\label{fig:ours_vs_supervised}
\end{figure*}

\begin{figure*}[h]
%\small
\footnotesize
\begin{tabularx}{\textwidth}{lXXXXX}
& \includegraphics[width=\linewidth]{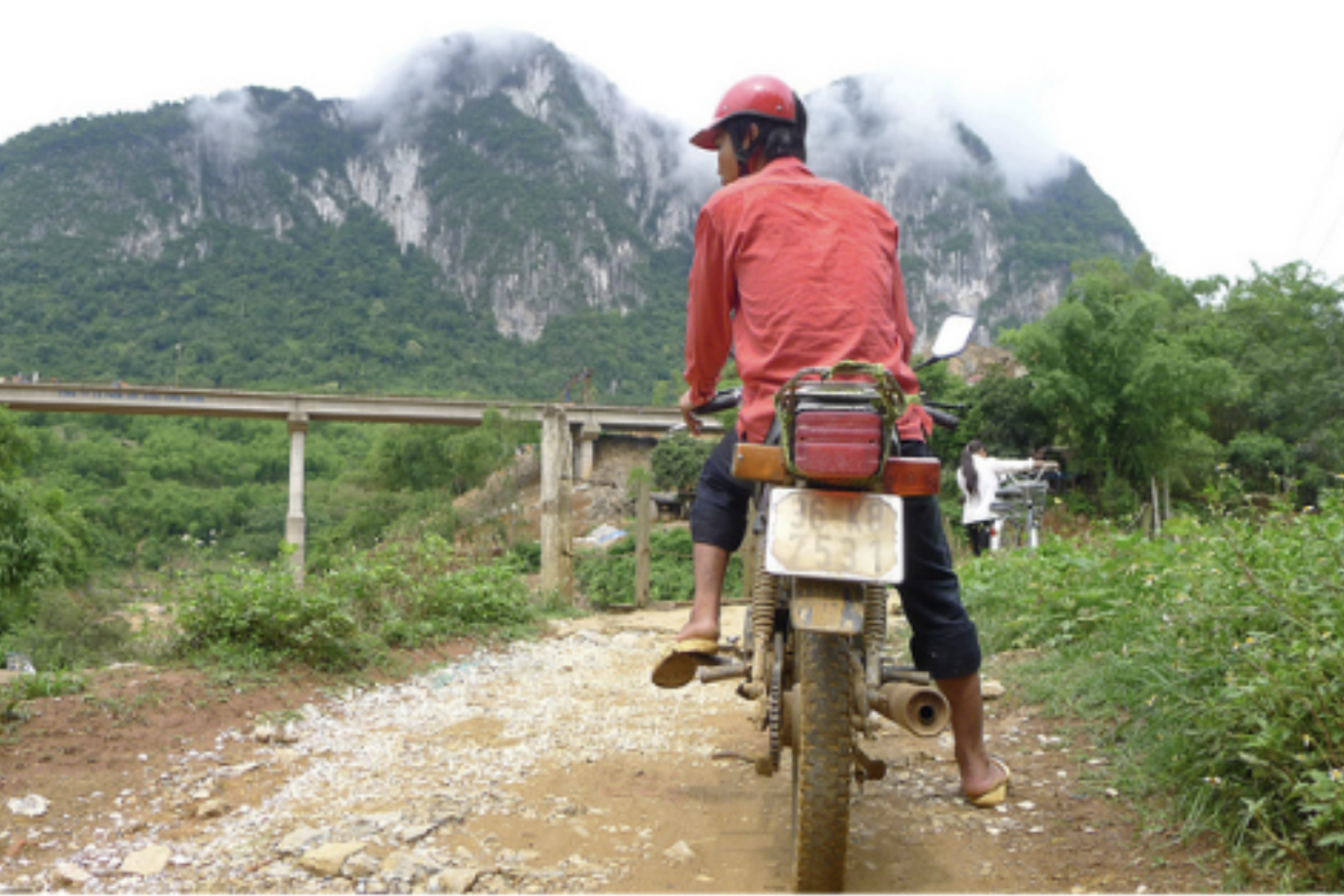} & \includegraphics[width=\linewidth]{figures/girl.png} & \includegraphics[width=\linewidth]{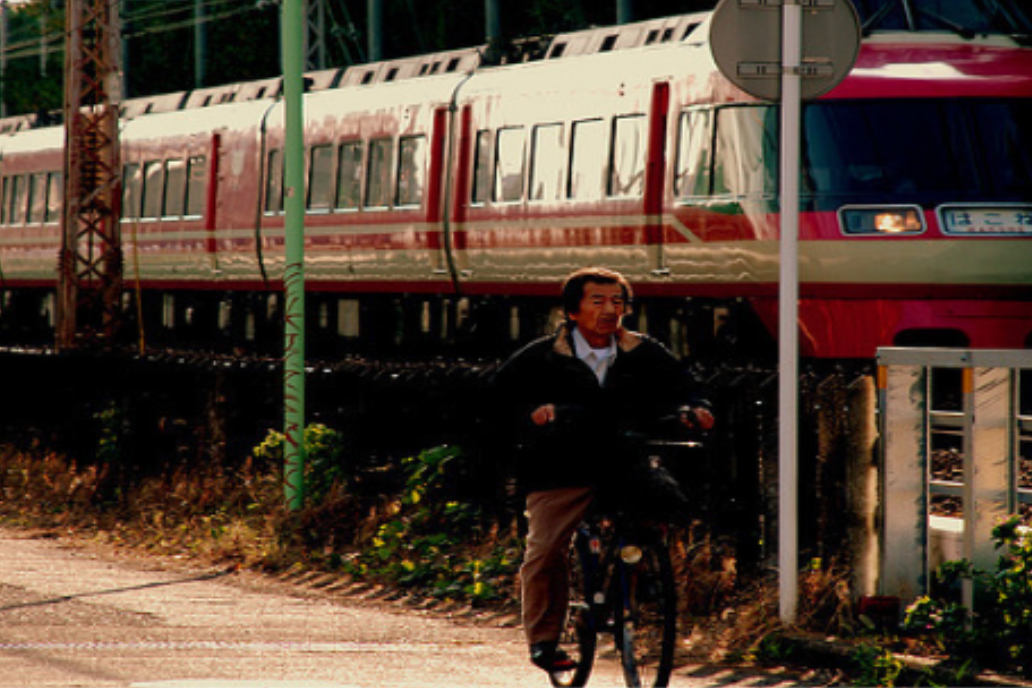} & \includegraphics[width=\linewidth]{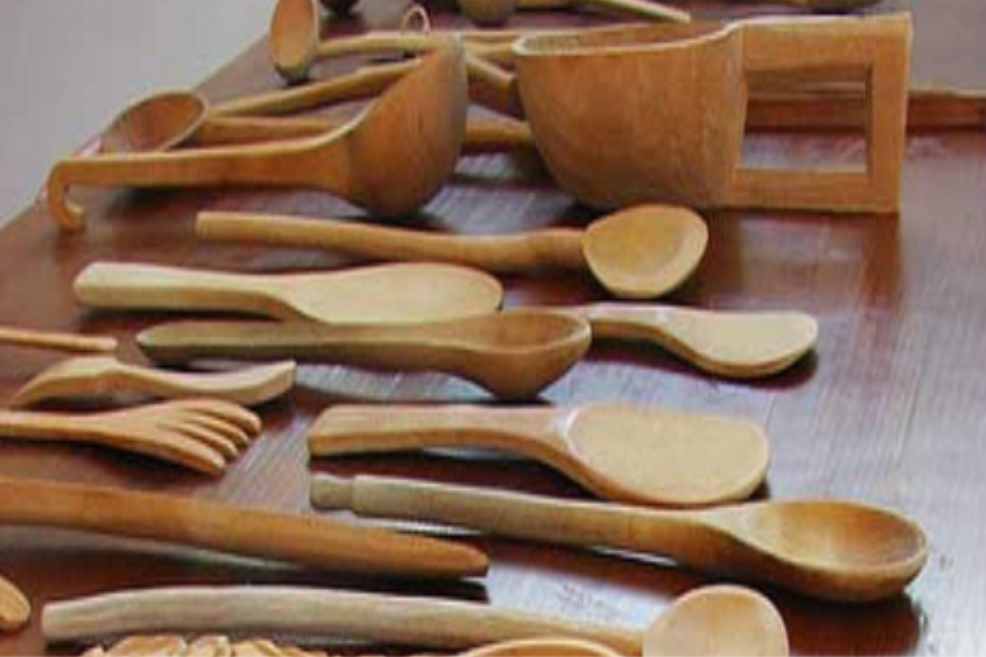} & \includegraphics[width=\linewidth]{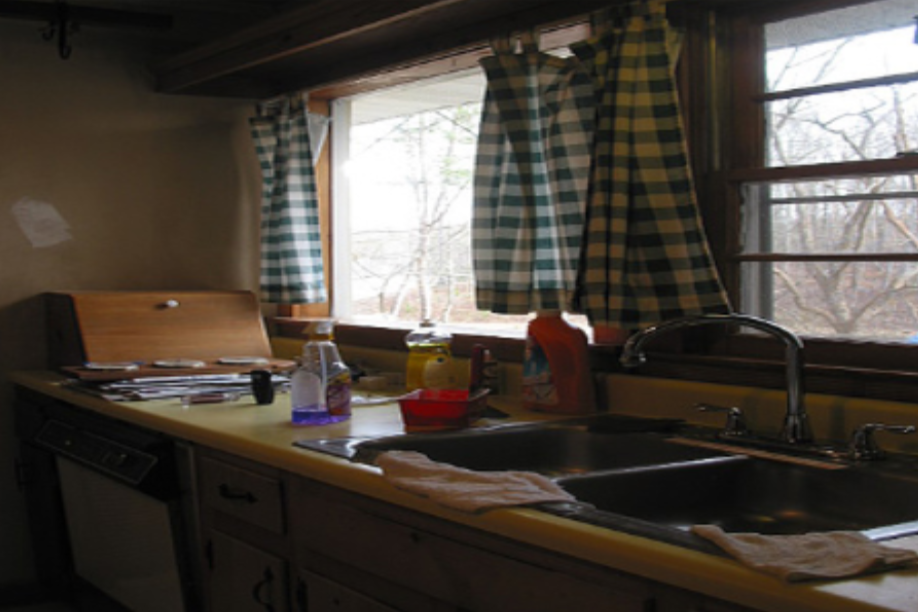} \\
\hline
Reference & 
A man with a red helmet on a small moped on a dirt road &
A young girl inhales with the intent of blowing out a candle &
A man on a bicycle riding next to a train &
A wooden cutting board topped with sliced up food & 
A kitchen is shown with a variety of items on the counters
\\ \hline
Oscar & 
A man riding a motorcycle down a dirt road ($0.22$) &
A woman sitting at a table with a plate of food ($0$) &
A woman riding a bike down a street next to a train ($0.02$) &
A woman sitting at a table with a plate of food ($0.42$) &
A kitchen with a sink, dishwasher and a window ($0.41$)
\\ \hline
%Supervised & 
%A man is riding a motorbike on a dirt road ($S_{\text{CLIP}} = 0.39$) &
%A young girl sitting at a table with a cup of cake ($S_{\text{CLIP}} = 0.01$) & 
%A man is standing next to a train ($S_{\text{CLIP}} = 0.01$) &
%A wooden table with a bunch of wood tools on it ($S_{\text{CLIP}} = 0.93$) & 
%A kitchen with a sink and a window ($S_{\text{CLIP}} = 0.19$)
%\\ \hline
Ours & 
A man riding a motorbike on a dirt mountain road
($0.37$) &
A woman is blowing out a candle on a piece of cake ($0.91$) & 
A man is riding a bike next to a train ($0.15$) &
A lot of wooden spoons on a wooden table ($1$) & 
A kitchen with a sink, a stove and a window ($0.31$)
\\
\end{tabularx}
%\vspace{\baselineskip}
\caption{Uncurated captions generated through both Oscar and our method for five images in the COCO-captions dataset. RefCompare scores are shown in parentheses.}
\label{fig:ours_vs_oscar}
\end{figure*}

In Figure~\ref{fig:ours_vs_supervised} and Figure~\ref{fig:ours_vs_oscar}, we show more examples of captions generated through our method, the supervised method, and Oscar.

\end{document}